\documentclass[runningheads]{llncs}
\usepackage{graphicx}
\usepackage{multirow}
\usepackage{booktabs}
\usepackage[misc]{ifsym}
\begin{document}
\title{Signet Ring Cell Detection With a Semi-supervised Learning Framework}

\author{Jiahui Li\inst{1} \and
Shuang Yang\inst{1} \and
Xiaodi Huang\inst{1} \and Qian Da \inst{2} \and Xiaoqun Yang \inst{2} \and \\
Zhiqiang Hu  \inst{1} \and Qi Duan \inst{1} \and Chaofu Wang \inst{2} \and Hongsheng Li \inst{3\ \textrm{\Letter}}}
\authorrunning{J.Li et al.}
%
\institute{SenseTime Research\\
\email{\{lijiahui,yangshuang1,huangxiaodi,huzhiqiang,duanqi\}@sensetime.com}\\ 
\and
Ruijin Hospital, Shanghai Jiao Tong University School of Medicine\\
\email{\{dq11848,yxq11964,wcf11956\}@rjh.com.cn}\\
\and
Chinese University of Hong Kong\\
\email{hsli@ee.cuhk.edu.hk}}
\maketitle              
\begin{abstract}
Signet ring cell carcinoma is a type of rare adenocarcinoma with poor prognosis.  Early detection leads to huge improvement of patients' survival rate. However, pathologists can only visually detect signet ring cells under the microscope. This procedure is not only laborious but also prone to omission.  An automatic and accurate signet ring cell detection solution is thus important but has not been investigated before.  In this paper, we take the first step to present a semi-supervised learning framework for the signet ring cell detection problem. Self-training is proposed to deal with the challenge of incomplete annotations, and cooperative-training is adapted to explore the unlabeled regions.  Combining the two techniques, our semi-supervised learning framework can make better use of both labeled and unlabeled data.  Experiments on large real clinical data demonstrate the effectiveness of our design.  Our framework achieves accurate signet ring cell detection and can be readily applied in the clinical trails. The dataset will be released soon to facilitate the development of the area.
%
\end{abstract}
\section{Introduction}
Signet ring cell carcinoma (SRCC) is an adenocarcinoma with a high degree of malignancy. SRCC is mostly found in stomach but could also spread to ovaries, lungs, breast, and other organs. The prognosis of SRCC is so poor that early detection leads to huge improvement of patients' survival rate. However, pathologists could only visually detect signet ring cells under the microscope, and/or confirm by immunohistochemistry. Manual detection is laborious and prone to omission, especially for scattered signet ring cells, while immunohistochemistry, which uses enzymes or other specific molecular markers to image antigens (protein) in abnormal cells, is expensive. After discussion with experienced pathologists and exploration of related medical background, we find that accurate cell edges have potential research value, such as the calculation of karyoplasmic ratio and the classification of degree of atypia. Therefore, an automatic and accurate signet ring cell detection solution is important and highly demanded.

\begin{figure}
\center
\includegraphics[width=0.7\textwidth]{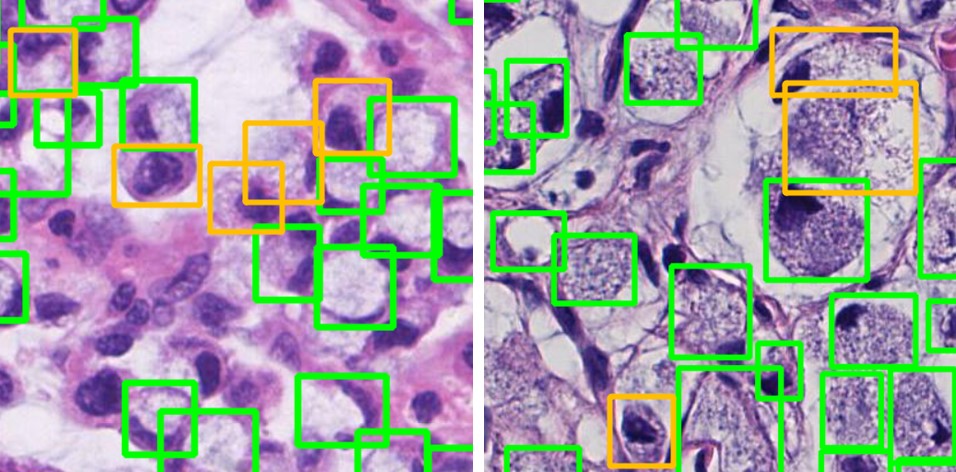}
\caption{Signet ring cells are overcrowded and of various appearances.  Cells in green rectangles are signet ring cells, which are the indicator of signet ring cell carcinoma. Cells in yellow rectangles are also signet ring cells but are missed by pathologists in crowd regions, during long time tedious annotation.} \label{intro-variaty}
\end{figure}

The unique appearance of the signet ring cells makes them completely different from other types of cells, characterized by a central optically clear, globoid droplet of cytoplasmic mucin with an eccentrically placed nucleus \cite{bosman2010classification}. To our best knowledge automatic signet ring cell detection has not been investigated before. In this paper, we take the first step to propose a signet ring cell detection framework based on semi-supervised learning.  We firstly collect and annotate a large real clinical signet ring cell dataset.  We collect 127 (21 positive + 106 negative) Hematoxylin and Eosin (H\&E) stained Whole Slide Images (WSIs) from 10 organs, including gallbladder, gastric mucosa, lymph, breast, ovary, pancreas, lung, urinary bladder, abdominal wall nodule and intestine. We select at least 3 regions of over $2,000\times2,000$ pixels from each positive/negative WSI, and annotate signet ring cell instances within the positive regions, in the form of the tight bounding box.  A total of $12,381$ signet ring cells are annotated.  However, overcrowded cells make it impossible to reach complete annotation, not to mention the occlusion and appearance variation, as shown in Fig. \ref{intro-variaty}.  As a result, pathologists can only guarantee that annotated cells are indeed signet ring cells; The opposite, that unannotated cells are not SRCC, is not necessarily true. The dataset will be released soon to facilitate the development of the area.

We propose a self-training method to deal with the challenge of incomplete annotations. We observe that some suspected areas predicted by the detector with high probabilities, are more likely to be real signet ring cells missed by pathologists during annotation. As a result, we combine those highly suspected areas with annotations to refine the labels, and take the refined labels to retrain the detector. The self-training strategy can be applied iteratively, with the labels further refined by the new highly suspected areas.

On the other hand, we propose a cooperative-training method to explore the unlabeled regions.  As mentioned above, we select 3 regions of over $2,000\times2,000$ pixels from each positive WSI for annotations.  In comparison, a single WSI has over $100,000\times100,000$ pixels in total, i.e., we only annotate a tiny fraction of the whole WSI. To make better use of the whole WSI, we select over 1,000 unlabeled regions from the positive WSIs, apply the inference process of two detectors with different backbones, take each others' highly suspected areas as labels and retrain the detectors with the augmented dataset. Two detectors are needed because detectors are prone to get stuck in their local minimum to general erroneous predictions for totally unlabeled regions and a cooperative way may help alleviate the situation. Similar to the self-training, cooperative-training strategy can also operate iteratively and two detectors benefits from each other in multiple rounds.

We perform extensive experiments on the collected dataset, and the results demonstrate the effectiveness of our design, where both the self-training and cooperative-training strategy deliver a significant and consistent improvement over the baseline.  Our semi-supervised learning framework achieves accurate signet ring cell detection and can be readily applied in clinical trails.

Our contributions are summarized as follows: we take the first step to investigate the signet ring cell detection problem and present a semi-supervised learning framework to tackle the problem. The self-training strategy is proposed to deal with the challenge of incomplete annotations, and the cooperative-training strategy is proposed to explore the unlabeled regions.  Combining the two techniques, the semi-supervised learning framework can make better use of both labeled and unlabeled data.  Experiments on large real clinical data demonstrate the effectiveness of our design.  

\subsubsection{Related Work} Several semi-supervised learning methods have been proposed and verified their effectiveness on natural images \cite{SemiSeg3,Semi5Selftrain2,luo2017deep,SemiSeg2,Seg4SelfTrain,SemiBox}. Papandreou et al. \cite{SemiSeg2} used box-level annotation to achieve similar performance with pixel-level annotation. Zhou et al. Luo et al. \cite{luo2017deep} organized a reconstruction head to convert segmentation output back to original images. Generally, all the previous methods focus on training one model with complex auxiliary branches. On the contrary, our semi-supervised learning framework can organize the multiple models to support each other and no extra auxiliary branch is needed. There have been lots of automatic methods for pathology, including nuclei segmentation \cite{zhang2015towards,zhang2015high,zach2009continuous} and specific object \cite{chen2016dcan}. However, automatic signet ring cell detection has not been investigated before to our best knowledge.

\begin{figure}[!htb]
\centering
\setlength{\abovecaptionskip}{0.cm}
\setlength{\belowcaptionskip}{.8cm}
\includegraphics[width=0.9\textwidth]{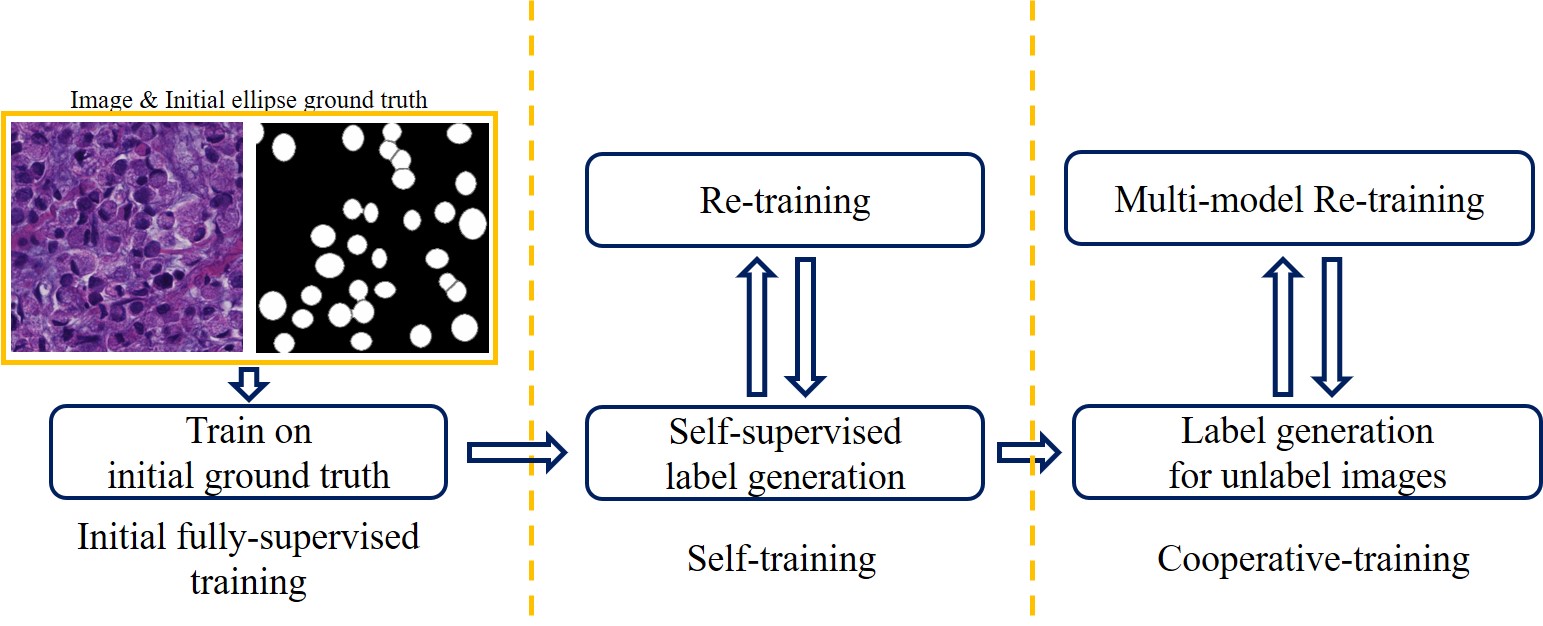}
\caption{Overview of our semi-supervised framework: Initial fully-supervised training, self-training, and cooperative-training.} \label{overview}
\end{figure}

\section{Semi-Supervised Learning Framework}

As shown in Fig. \ref{overview}, our semi-supervised learning framework consists of three steps: initial fully-supervised training, self-training and cooperative-training. 

\begin{figure}[!htb]
\centering
\includegraphics[width=0.7\textwidth]{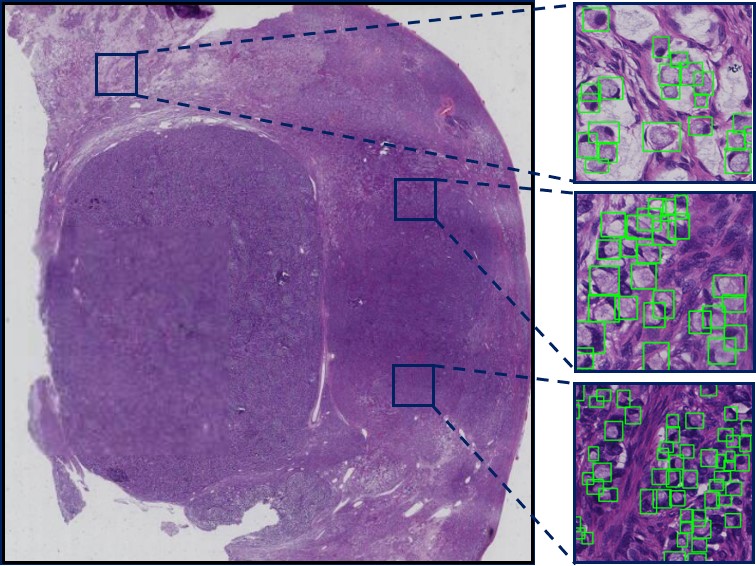}
\caption{3 regions of over $2,000\times2,000$ pixels are randomly cropped from each WSI and bounding boxes for high confident independent signet ring cells are annotated.} \label{dataset-description}
\end{figure}

\subsection{Dataset} Our dataset is collected from several highly ranked hospitals, and consists of H\&E stained images captured at 40$\times$ magnification. Containing 127 (21 positive + 106 negative) whole slide images (WSIs), this dataset covers a large number of patients and images from 10 organs, including gallbladder, gastric mucosa, lymph, breast, ovary, pancreas, lung, urinary bladder, abdominal wall nodule and intestine. For each WSI, pathologists carefully annotate each independent signet ring cell by tight bounding box in at least 3 regions of over $2,000\times2,000$ pixels, as shown in Fig. \ref{dataset-description}. For each region one pathologist provide annotation verified by one senior pathology. Thus each labeled cells are indeed signet ring cells. A total of 74 regions are annotated for 21 tumor WSIs. Bounding box is a convenient way to conduct annotations for thousands of cells, which saves time and offers instance analysis such as cell counting. We also randomly crop 320 regions of the same size from the other 106 negative WSIs, which are either healthy or infected by other types of cancer. As a result, we collect 74 annotated positive regions from 21 positive WSIs of 10 organs and 320 negative regions from 106 negative WSIs, with a total of $12,381$ signet ring cells annotated. 

\subsection{Initial fully-supervised training}
\begin{figure}[!htbp]
\center
\includegraphics[width=0.85\textwidth]{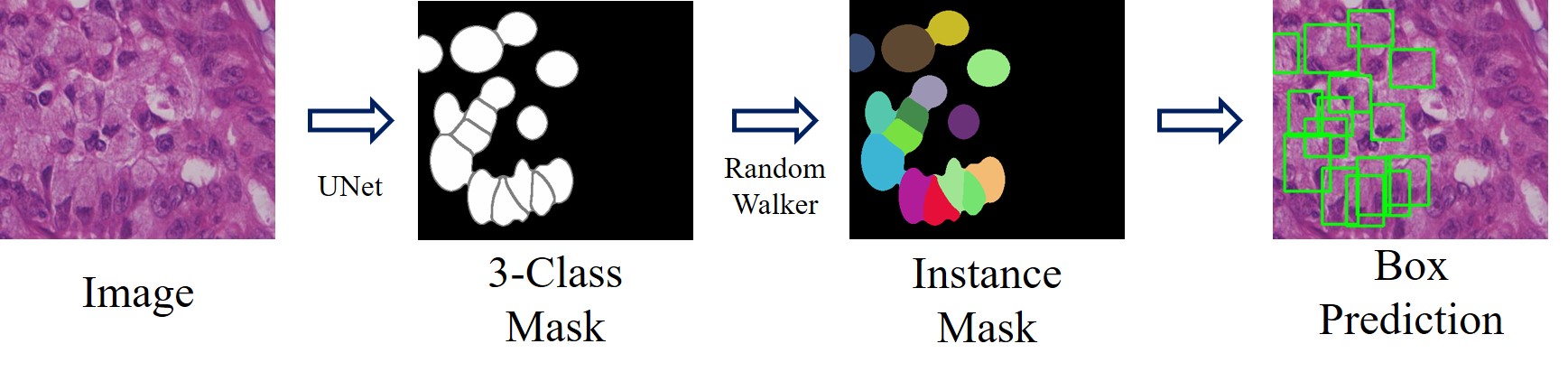}
\caption{Signet ring cell detector: image $\to$ 3-class mask $\to$ cell instance mask $\to$ cell box prediction.} \label{SRCDetector}
\end{figure}

Firstly we will introduce the initial fully-supervised training with the original ground truth annotations, which is the first stage of our signet ring cell detector (SRCDetecor).  This also serves as the baseline of our proposed semi-supervised learning framework. Unlike other common RCNN-based detection frameworks \cite{ren2015faster,redmon2016you}, We propose a bottom-up method to directly predict cell instance mask, then derive boxes for each instance, as shown in Fig. \ref{SRCDetector}, our proposed SRCDetector adopts a UNet \cite{MedicalUnet} to perform 3-class segmentation, i.e., classifying images into background, cell edges, and cell inner regions. Then use Random Walker \cite{RandomWalker} to transform the obtained 3-class mask to cell instance mask, where cell inner regions are seeds and cell edges are undetermined regions. Finally we extract bounding box of each instance as final box prediction. To train the UNet, we first extract the inscribed ellipse of each annotated bounding box and take the inner region/edge of the ellipse as ground truth inner regions and edges.  Our loss function is the summation of cross entropy loss and Intersection-Over-Union (IOU) loss. In (1) and (2), $y_i$ and $p_i$ are targets and predictions for pixel $i$ respectively. E is the edge of the cell and IR is the inner region of the cell. 

\begin{eqnarray}
l_{\mathrm{CE}} & = & \sum_{S \in \{\mathrm{E, IR}\}} \sum_{i \in S} y_i \log(p_i) + (1-y_i) \log(1-p_i)\\
l_{\mathrm{IOU}} & = & \sum_{S \in \{\mathrm{E, IR}\}} 1 - \frac{\sum_{i \in S} y_i p_i}{\sum_{i \in S} y_i + \sum_{i \in S} p_i - \sum_{i \in S} y_i p_i}\\
l & = &  l_{\mathrm{CE}} +  l_{\mathrm{IOU}}
\end{eqnarray}

\subsection{Self-training} 
\begin{figure}
\centering
\includegraphics[width=0.85\textwidth]{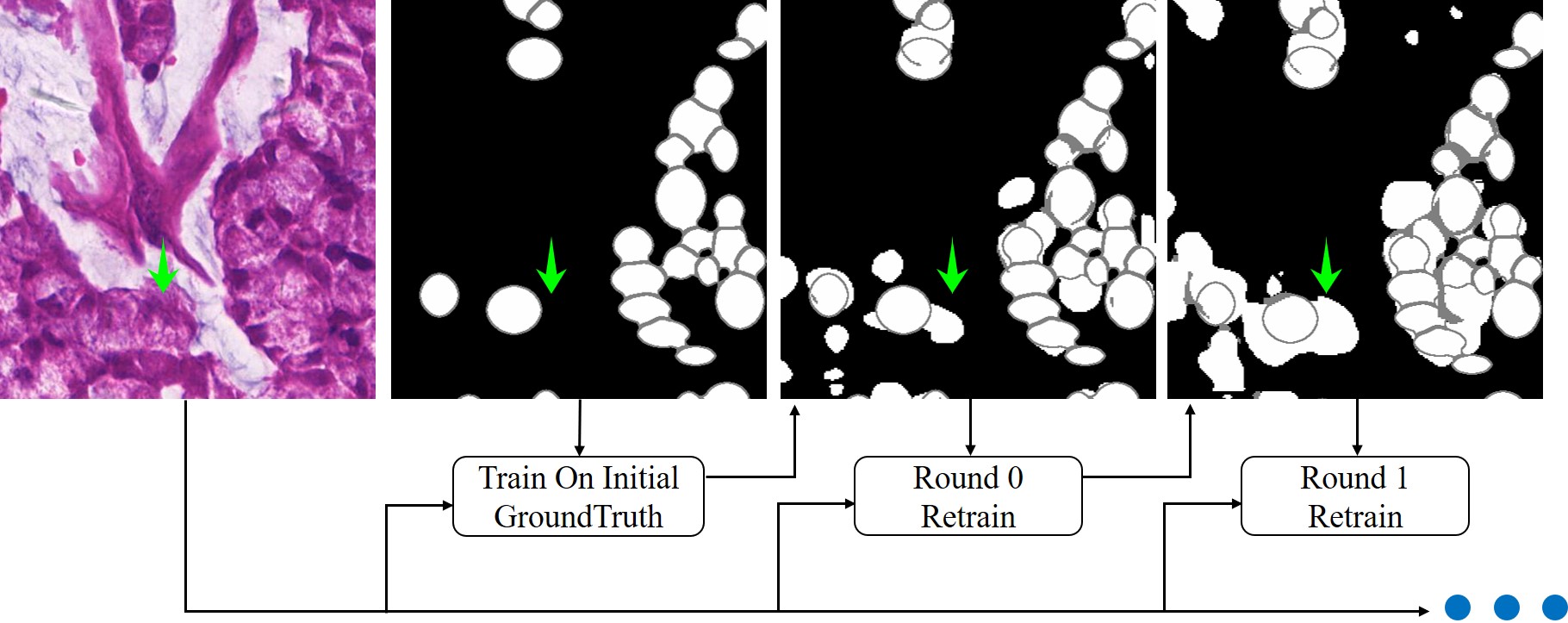}
\caption{Pipeline of our self-training strategy to self-correct imperfectly annotated images. The next-round model is trained on annotations from the previous round, and iteratively adjusts annotations towards higher quality. Initially we draw inscribed ellipse in each rectangle as ground truth inner region. The gray regions are edge mask and the white regions are the inner regions. The green arrow points to a growing correct prediction.} \label{self-training}
\end{figure}

After training with ellipse mask obtained from the original annotation boxes, our proposed SRCDetector would generate some suspected positive areas in the training images. It is impossible to ask pathologists to annotate all the signet ring cell for overcrowdness and appearance variation. Confirmed by senior pathologists, those suspected positive areas with high confidence are indeed signet ring cells for most cases, and could be combined with the initial ellipses mask to create the refined labels for retraining the 3-class segmentation task again. As shown in Fig. \ref{self-training}, the pipeline of self-training is an iterative process, whose main step is to use the previously trained model to generate new ground truth and refine cell edges to train next model. The 3-class mask merges with annotations by following steps: 1, use the union inner region of the prediction of the current round and the inscribed ellipse of the initial ground truth as the inner region mask. 2, use the union edge of the prediction cell in the current round and the inscribed ellipse of the initial ground truth as the next edge, overwrite pixels which are regarded as inner region in step 1. 3, use the rest pixels as cell background. These steps could be repeated for several times, until a well-annotated pixel-wise mask is generated, and should be stopped if the newly predicted mask stops to grow. Generally speaking, the self-training strategy gradually generates more annotations, which are mostly missed by junior pathologists and can improve the quality of annotations by the iterative refinement. However, 74 annotated images are not enough to train a robust model, we turn to generate more available labels in unknown areas with cooperative-training strategy. 

\begin{figure}
\centering
\includegraphics[width=0.85\textwidth]{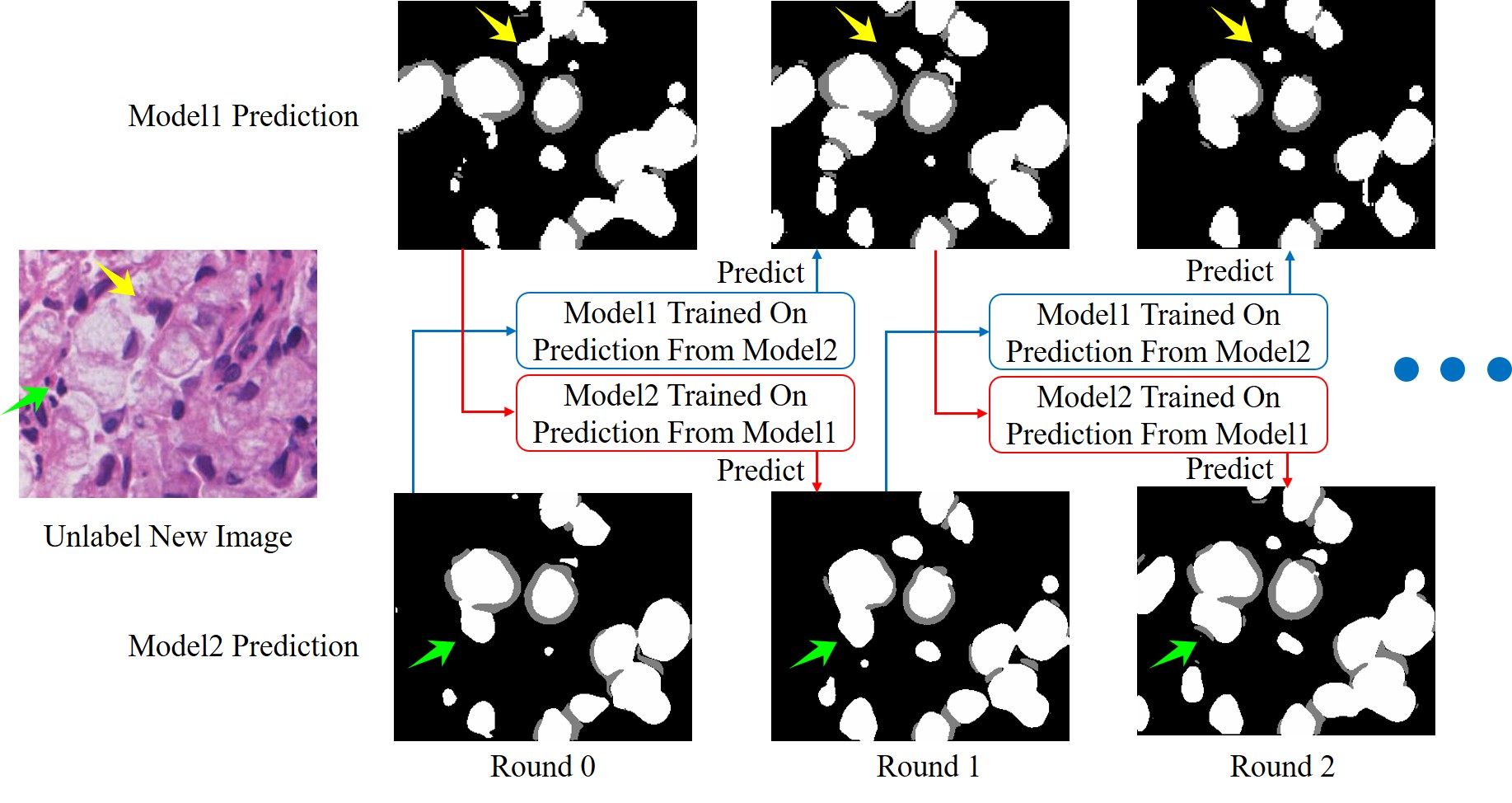}
\caption{Pipeline of the cooperative-training strategy. For new unlabeled images, each model is trained on predictions from the other model, so that we can reduce the possibility that one model get stuck in its local minimum , and allow the two models to support each others. In this way we will gradually obtain higher annotation quality on unlabeled images without any manual interventions. The yellow arrow points to a wrong prediction that gradually shrinks. The green arrow points to a growing correct prediction. } \label{cooperative-training}
\end{figure}

\subsection{Cooperative-training} As each WSI is an image with ultra high resolution of over $100,000\times100,000$ pixels at 40$\times$ magnification, in which only 3 randomly selected regions of over $2,000\times2,000$ pixels are manually labeled. The rest regions are also expected to contain possible positives that may provide additional training samples for our SRCDetector. The self-training strategy has an obvious drawback that it might get stuck in local minimum to general erroneous predictions. To alleviate this problem, we use two SRCDetectors with different backbones, and train with generated labels from each other to reduce self amplify errors. Firstly we use the self-training strategy to train two models with different backbones, and do inference on unlabeled images. Secondly each model is trained with generated labels from the other model's predictions from the previous model. Again in the next round, two models do inference on unlabeled images and retrain on each others' predictions. Annotated images are mixed with unlabeled images to train during Cooperative-Training. As shown in Fig. \ref{cooperative-training}, two models are organized to support each other in an interative way: next-round model is trained with predictions from the other model in the previous round, and stop until no more significant changes in unlabeled images.

Models for cooperative-training shall be able to provide complementary information for each other. We achieve this by using two models with different structures and different training data for two models. In this problem, two models are trained on same images but different annotations predicted from each other, which can be regarded as different training data.



\section{Experiment}
\subsection{Evaluation}
As mentioned above, this is an imperfect annotated dataset and we can only guarantee that the labeled cells are confidently true. Hence for detection evaluation \cite{VOC2012}, recall is still reasonable to measure while precision becomes meaningless. Due to the incomplete annotations, we add three criteria for evaluation. Firstly, instead of precision, normal region false positives (FPs) is taken into consideration, which means the average number of wrong prediction boxes, which are definitely false positives, in negative images. Secondly, in order to comprehensively consider recall and normal region false positives, we define the FROC \cite{FROC}: By adjusting confidence threshold, when the number of normal region false positives is 1, 2, 4, 8, 16, 32, the FROC is the average of relevant recall at those confidence threshold. Thirdly, besides the common instance-level recall, we propose another criteria called collective-level recall, which means we draw a big mask with the union of all the prediction boxes, then if any annotation box's intersection area divided by this box's area is larger than some threshold, we regard this box as being detected. Collective-level recall is an instance unawareness criteria and is the upper bound of instance-level recall. Comparing the performance difference between instance-level recall and collective-level recall, we can learn whether model performs poorly at SRC region separation or SRC detection. For FROC, we only consider instance-level FROC. In conclusion, four criteria shall be considered: collective-level recall, instance-level recall, normal region false positives, instance-level FROC.


For pathological images, we can get images under same distribution by cropping out different regions from the same WSI. However, images from different WSIs and organs differ from each other substantially. The total available images are limited, thus cross validation shall be performed under an easy mode and a hard mode. In the easy mode, we can assume that test data and train data are of the same distribution, because images from a same WSI are assigned randomly to different folds, allowing models to see similar images both in train and test data. This man-made distribution makes it possible for us to understand what the best performance will be if our dataset is big enough.  In hard mode, images in different folds come from different WSIs and organs. Because signet ring cell can appear in other kinds of organs, which may not be collected in our dataset. The hard mode can help us evaluate model's capability when dealing with unknown organs.

\subsection{Implementation Details}
The SRCDetector is implemented with PyTorch 0.4, using the Adam optimizer and 0.001 as learning rate. During training, we randomly crop images of size $512\times512$ pixels with the batch size of 16 as input, and use sliding windows of size $1,024\times1,024$ pixels during inference. For cooperative-training we use UNet with two different backbones: ResNet \cite{Resnet} and Deep Layer Aggregation (DLA) \cite{DLA}. We use 34-layer models for both backbones. Our SRCDetector is trained with original ground truth by 50 epochs and then in iterating self-training with extra annotations for 10 epochs in each round. Taking probability larger than 0.7 to identify new positive regions, the self-training stops at round 5 in our experiments. Similarly, the two models in cooperative-training are trained on annotation from the self-training for 50 epochs and for 10 epochs in each round, takes probability larger than 0.33 to identify new positive regions and stops at round 2. The entire procedure requires 8 hours for fully supervised training, 30 hours of 5 rounds for self-training and 24 hours of 2 rounds for  cooperative-training. We only use 2 models for cooperative-training because we observe there is no obvious benefit if 3 or more models are utilized, which do not increase the final accuracy, and also decrease the computational efficiency. During inference, DLA is utilized which usually performs 2\% better than Resnet. Models from round 5 self-training and round 2 cooperative-training is used to conduct inference on test data. In all training procedures, positive or negative images are duplicated for several times to achieve data balance. In our experiments, three kinds of SRCDetector are compared both in easy and hard modes, with the same structures and different training data as shown below:

{\bfseries Fully-supervised training.} Train on 74 images + initial ellipse annotations, and 320 negative images.

{\bfseries Self-training.} Train on 74 images + fixed annotations from self-training, and 320 negative images.

{\bfseries Self-training-extra.} Train on 74 images + fixed annotations from self-training, 1236 unlabeled images + generated annotations from self-training, and 320 negative images.

{\bfseries Cooperative-training.} Train on 74 images + fixed annotations from self-training, 1236 unlabeled images + generated annotations from cooperative-training, and 320 negative images.

During evaluation, both in the easy and hard mode, only 74 images' initial manual annotations are considered for recall evaluation. To calculate collective-level recall, instance-level recall, normal region false positives and instance-level FROC, we perform 3-folds cross-validation to fully evaluate SRCDetector on 74 positive and 320 negative images. Predict and groundtruch boxes are match if their IOU is greater than 0.3.

\subsection{Results and Discussion}
As shown in Table. \ref{performance}, compared with fully-supervised training on initial ground truth, which is the baseline, both two steps of our semi-supervised learning framework introduce obvious improvements.

\begin{table}[!htbp]
\caption{Cross validation performance comparison under different modes and data utilities.  Col Recall, Ins Recall, Nor FPs and Ins FROC are short for collective-level recall, instance-level recall, normal region false positives and instance-level FROC, while FT, ST , ST-Ex and CT are short for fully-supervised training, self-training, self-training-extra and cooperative-training, respectively.}\label{performance}
\centering
{\renewcommand{\arraystretch}{2}
\begin{tabular}{rcccccccc}
\toprule
\multirow{2}*{~~Criteria~~~} & \multicolumn{4}{c}{Easy Mode} & \multicolumn{4}{c}{Hard Mode}\\
\cmidrule(lr){2-5} \cmidrule(lr){6-9}
&~~~FT~~~&~~~ST~~~&~~~ST-Ex~~~&~~~CT~~~&~~~FT~~~&~~~ST~~~&~~~ST-Ex~~~&~~~CT~~~~~\\
\midrule
~~~Col Recall~~~& 0.626&0.869&0.841&0.881& 0.497&0.693&0.670&0.827\\
~~~Ins Recall~~~& 0.462&0.673&0.638&0.705& 0.325&0.521&0.505&0.658\\
~~~Nor FPs~~~& 0.446&2.29&5.22&1.45& 0.202&1.18&4.62&0.943\\
~~~Ins FROC~~~& 0.462&0.617&0.562&\textbf{0.692}& 0.325&0.517&0.451&\textbf{0.657}\\
\bottomrule
\end{tabular}
}
\end{table}

According to the comparison between baseline and self-training, the major contribution of self-training is to improve the absolute performance: 0.16 improvement on instance-level FROC in the easy mode and 0.18 improvement on instance-level FROC in the hard mode. During training on initial ground truth, we find that in train data, models focus much on suspected areas which are indeed signet ring cells that are not annotated. These areas are treated as negative samples in training phase, such false negatives are harmful for training the detector. After visually confirmed by senior pathologists, most conflicts between prediction and ground truth come from unlabeled signet ring cells. Self-training could suppress the ratio of noisy annotations to provide annotations of higher quality.

According to the comparison between self-training and cooperative-training, the major contribution of cooperative-training is to narrow the performance gap between the easy mode and the hard mode and introduce small improvement in the easy mode, with a 0.07 improvement on instance-level FROC. In general, we observe that training on extra 50 regions in the same WSI could only slightly augment the performance than on annotated 2 regions. Therefore, to improve the performance in the easy mode, quality improvement of annotations shall be of highest priority in the future. In the hard mode cooperative-training still introduces obvious improvement, with a 0.14 improvement on instance-level FROC, meaning that in self-training there exists large data gap between different WSIs and organs, and this gap can be narrowed by using extra annotations from same WSIs in train data. Even with noisy annotations generated by cooperative-training, models can learn much common morphological varieties across different WSIs and organs. However, we can still observe a few images in hard mode with only 23\% recall, which demonstrate that signet ring cells in different organs do show different appearance.

\begin{figure}[!htbp]
\centering
\includegraphics[width=0.8\textwidth]{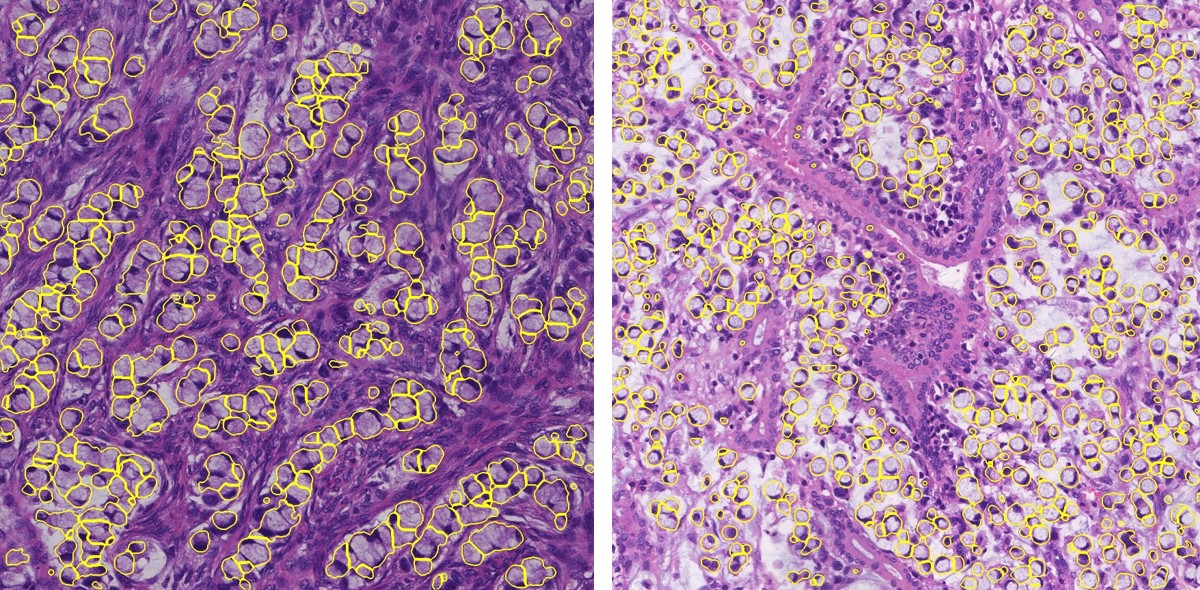}
\caption{Example detection results in test images of over $2,000\times2,000$ pixels. Our semi-supervised learning framework can obtain cell edges of each signet ring cell as shown in yellow polygon. } \label{detection-visualization}
\end{figure}

Comparing instance-level recall with collective-level recall, we find that model has found 82.7\% signet ring cells in the hard mode, but has difficulties in separating neighbor signet ring cells. This will harm the accuracy of signet ring cell counting, as shown in Fig. \ref{detection-visualization}.

Comparing self-training with self-training-extra, unlabeled images with self-generated annotations can harm the performance, especially in easy mode  where the false positive rises to 5.22. Visually we find that there are too many neutrophils, plasma cells, and gland cells in unlabeled images being annotated. However the three types of cell are rarely annotated as false positives in 74 partially labeled images. While in cooperative-training these three types of specific false positives are not severely amplified. Therefore self-training will indeed amplify the errors while cooperative-training does not show this phenomenon.

\section{Conclusion}
In this paper we propose a semi-supervised learning framework to make better use of collected data with relative small amount of annotation cost. Semi-supervised learning framework not only improves the quality of annotation but also makes better use of extra unlabeled images. We verify the improvement by collecting a multi-organ signet ring cell dataset and propose a series of new evaluation metrics for imperfect annotated data. In the future we will try to understand why specific types of false positive are not amplified in cooperative-training. The proposed dataset will be released soon to facilitate the development of the area.


%
%
%
\bibliographystyle{splncs04}

\begin{thebibliography}{10}
\providecommand{\url}[1]{\texttt{#1}}
\providecommand{\urlprefix}{URL }
\providecommand{\doi}[1]{https://doi.org/#1}

\bibitem{FROC}
Bandos, A.I., Rockette, H.E., Song, T., Gur, D.: Area under the free-response
  roc curve (froc) and a related summary index. Biometrics  \textbf{65}(1),
  247--256 (2009)

\bibitem{bosman2010classification}
Bosman, F.T., Carneiro, F., Hruban, R.H., Theise, N.D., et~al.: WHO
  classification of tumours of the digestive system., pp. 52--53. No. Ed. 4,
  World Health Organization (2010)

\bibitem{chen2016dcan}
Chen, H., Qi, X., Yu, L., Heng, P.A.: Dcan: deep contour-aware networks for
  accurate gland segmentation. In: Proceedings of the IEEE conference on
  Computer Vision and Pattern Recognition. pp. 2487--2496 (2016)

\bibitem{VOC2012}
Everingham, M., Van~Gool, L., Williams, C.K.I., Winn, J., Zisserman, A.: The
  pascal visual object classes (voc) challenge. International Journal of
  Computer Vision  \textbf{88}(2),  303--338 (Jun 2010)

\bibitem{RandomWalker}
Grady, L.: Random walks for image segmentation. IEEE transactions on pattern
  analysis and machine intelligence  \textbf{28}(11),  1768--1783 (2006)

\bibitem{Resnet}
He, K., Zhang, X., Ren, S., Sun, J.: Deep residual learning for image
  recognition. In: Proceedings of the IEEE conference on computer vision and
  pattern recognition. pp. 770--778 (2016)

\bibitem{Semi5Selftrain2}
Hosang, A.K.R.B.J., Schiele, M.H.B.: Weakly supervised semantic labelling and
  instance segmentation. arXiv preprint arXiv:1603.07485  (2016)

\bibitem{SemiSeg3}
Hu, R., Doll{\'a}r, P., He, K., Darrell, T., Girshick, R.: Learning to segment
  every thing. In: Proceedings of the IEEE Conference on Computer Vision and
  Pattern Recognition. pp. 4233--4241 (2018)

\bibitem{luo2017deep}
Luo, P., Wang, G., Lin, L., Wang, X.: Deep dual learning for semantic image
  segmentation. In: Proceedings of the IEEE Conference on Computer Vision and
  Pattern Recognition, Honolulu, HI, USA. pp. 21--26 (2017)

\bibitem{SemiSeg2}
Papandreou, G., Chen, L., Murphy, K., Yuille, A.L.: Weakly- and semi-supervised
  learning of a {DCNN} for semantic image segmentation. CoRR
  \textbf{abs/1502.02734} (2015), \url{http://arxiv.org/abs/1502.02734}

\bibitem{Seg4SelfTrain}
Rajchl, M., Lee, M.C., Oktay, O., Kamnitsas, K., Passerat-Palmbach, J., Bai,
  W., Damodaram, M., Rutherford, M.A., Hajnal, J.V., Kainz, B., et~al.:
  Deepcut: Object segmentation from bounding box annotations using
  convolutional neural networks. IEEE transactions on medical imaging
  \textbf{36}(2),  674--683 (2017)

\bibitem{redmon2016you}
Redmon, J., Divvala, S., Girshick, R., Farhadi, A.: You only look once:
  Unified, real-time object detection. In: Proceedings of the IEEE conference
  on computer vision and pattern recognition. pp. 779--788 (2016)

\bibitem{ren2015faster}
Ren, S., He, K., Girshick, R., Sun, J.: Faster r-cnn: Towards real-time object
  detection with region proposal networks. In: Advances in neural information
  processing systems. pp. 91--99 (2015)

\bibitem{MedicalUnet}
Ronneberger, O., Fischer, P., Brox, T.: U-net: Convolutional networks for
  biomedical image segmentation. In: International Conference on Medical image
  computing and computer-assisted intervention. pp. 234--241. Springer (2015)

\bibitem{DLA}
Yu, F., Wang, D., Shelhamer, E., Darrell, T.: Deep layer aggregation. In:
  Proceedings of the IEEE Conference on Computer Vision and Pattern
  Recognition. pp. 2403--2412 (2018)

\bibitem{zach2009continuous}
Zach, C., Niethammer, M., Frahm, J.M.: Continuous maximal flows and wulff
  shapes: Application to mrfs. In: Computer Vision and Pattern Recognition,
  2009. CVPR 2009. IEEE Conference on. pp. 1911--1918. IEEE (2009)

\bibitem{zhang2015towards}
Zhang, X., Liu, W., Dundar, M., Badve, S., Zhang, S.: Towards large-scale
  histopathological image analysis: Hashing-based image retrieval. IEEE
  Transactions on Medical Imaging  \textbf{34}(2),  496--506 (2015)

\bibitem{zhang2015high}
Zhang, X., Xing, F., Su, H., Yang, L., Zhang, S.: High-throughput
  histopathological image analysis via robust cell segmentation and hashing.
  Medical image analysis  \textbf{26}(1),  306--315 (2015)

\bibitem{SemiBox}
Zhao, X., Liang, S., Wei, Y.: Pseudo mask augmented object detection. In:
  Proceedings of the IEEE Conference on Computer Vision and Pattern
  Recognition. pp. 4061--4070 (2018)

\end{thebibliography}
%

\end{document}